\documentclass{article}
\usepackage{spconf,amsmath,graphicx,amssymb}
\usepackage{bm}
\usepackage{graphicx}
\usepackage{subfigure}
\usepackage{verbatim}

\title{Excavating RoI Attention for Underwater Object Detection}
\name{Xutao Liang\textsuperscript{1}, Pinhao Song\textsuperscript{2*} \thanks{This work is suported by Science and Technology Development Fund of Macau (0008/2019/A1, 0010/2019/AFJ, 0025/2019/AKP)}\thanks{*Corresponding author: pinhaosong@pku.edu.cn.}}
\address{\textsuperscript{1}State Key Laboratory of Lunar and Planetary Sciences, Macau University of Science and Technology\\
  \textsuperscript{2}Shenzhen Graduate School, Peking University
}

\begin{document}

\maketitle

\begin{abstract}
  Self-attention is one of the most successful designs in deep learning, which calculates the similarity of different tokens and reconstructs the feature based on the attention matrix. Originally designed for NLP, self-attention is also popular in computer vision, and can be categorized into pixel-level attention and patch-level attention. In object detection, RoI features can be seen as patches from base feature maps. This paper aims to apply the attention module to RoI features to improve performance. Instead of employing an original self-attention module, we choose the external attention module, a modified self-attention with reduced parameters. With the proposed double head structure and the Positional Encoding module, our method can achieve promising performance in object detection. The comprehensive experiments show that it achieves promising performance, especially in the underwater object detection dataset. The code will be avaiable in: https://github.com/zsyasd/Excavating-RoI-Attention-for-Underwater-Object-Detection

\end{abstract}


\section{Introduction}
Transformer has become the dominant architecture of design model in natural language processing (NLP), such as BERT \cite{Bert}, GPT serise, \cite{gpt1,gpt2} and RoBERTa \cite{Roberta}. They achieve fantastic performance on various tasks, such as text classification, machine translation, and question answering. Self-attention is the core module in Transformer \cite{Transformer}, which can excavate the relation between words and reconstruct word embedding. Because the self-attention module calculates the similarity between every two words, it can overcome the long-range dependencies problem of the RNN-based models. 

Non-local Neural Networks \cite{nl} brings the self-attention mechanism from NLP to Computer Vision (CV). Non-local block calculates the similarity between pixels in the feature maps and reconstructs the pixel features based on the attention matrix. However, the number if pixels is much larger than that of words. Therefore, several works are needed to design a modified non-local block to reduce the calculation complexity \cite{crisscross,standalone,ea}.

ViT \cite{ViT} proposes a much more radical idea: building a pure Transformer-based backbone without convolution neural networks. It separates the whole image into 16 $\times$ 16 patches, flattens and regards them as tokens. Using Transformer Encoder only and pre-trained on JFT-300M, ViT obtains superior performance on various image recognition datasets (ImageNet, CIFAR-100, VTAB, etc.). After that, many researchers join this gold rush of the backbone design based on Transformer, and propose many valuable methods. Most of them are based on the inter-patch attention \cite{swin,pvt,deit}.

In this paper, we adopt the idea of patch-level attention on object detection. In two-stage object detection like Faster R-CNN \cite{ren2015faster}, base features from the backbone are fed into Region Proposal Network (RPN) to obtain a series of proposals. With the proposals and base features, RoIPooling/RoIAlign crops the features and then resize them to a certain size, which is denoted as Region of Interest (RoI). RoI can be seen as one patch of the feature maps. Based on existing works of patch-level attention, it is possible to capture the RoI-level relation with the attention module to improve the performance.

Therefore, we propose a new detector named RoIAttn. First, we design a RoI attention module, which applies external attention to RoI features \cite{ea}. Second, separated heads for classification and regression respectively are used to improve the performance. Third, since it is difficult for convolution to regress the coordinates, positional encoding is applied on the regression branch. Comprehensive experiments on UTDAC2020 datasets show the effectiveness of our method. The experiment on PASCAL VOC dataset also show a competitive performance.


\begin{figure*}[t]
  \centering
  \includegraphics[width=15cm]{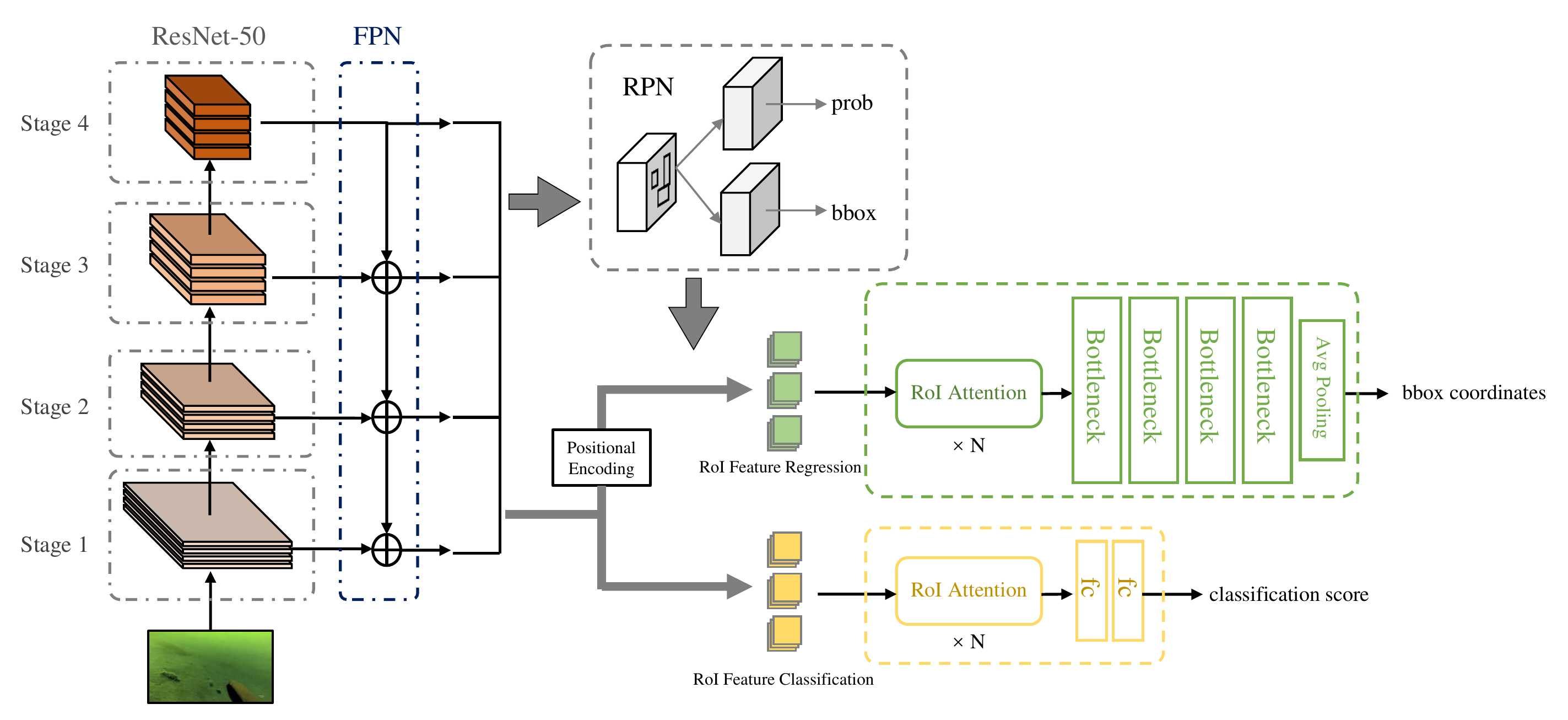} 
  \caption{The overview of the proposed RoIAttn. The green box denotes the regression branch, while the yellow box denotes the classification branch. The parameters of RoI attention modules in two branches are shared.}
  \label{fig:pipeline}
\end{figure*}

\section{Related Works}
\subsection{Object detection}
Existing object detection can be categorized into two mainstreams patterns: two-stage detectors and one-stage detectors. For two-stage detectors, the basic idea is to transform the detection task into classification problem. In the first stage, the RPN \cite{ren2015faster} proposes candidate object bounding boxes, and RoIPooling/RoIAlign is leveraged to crop the features from backbone. In the second stage, R-CNN head realizes classification and regression tasks of all objects. Some works are focus on modifying the R-CNN head to improve the performance \cite{sun2021sparse, doublehead, cai2018cascade,zhang2020dynamic}. One-stage detectors abandon the RPN and RoIPooling/RoIAlign, and directly regress the coordinates of bounding boxes and classes of the objects \cite{liu2016ssd, retinanet,fsaf}. 

\subsection{Attention and Transformer in computer vision.}
Since the self-attention module \cite{Transformer} is proposed and applied in NLP, researchers started to transfer this module to CV. The non-local block \cite{nl} is the self-attention module in CV, and it attempts to model long-range dependencies in both space and time. However, the non-local block suffers from great computational and memory costs. Criss-cross attention \cite{crisscross} focuses only on criss-cross relation on the feature maps, which reduces the computational complexity. Stand-alone self-attention \cite{standalone} performs local self-attention on kernel size like CNN filter. External Attention \cite{ea} analyzes the formulation of self-attention, and proposes a simplified attention module with only two linear layers.

Recently, research works have been focused on reforming the backbone with Transformer architecture. Vision Transformer \cite{ViT} employs a pure Transformer structure, achieving astonishing results in various image recognition benchmarks. DeiT \cite{deit} further uses a distillation approach to train ViT. Pyramid Vision Transformer (PVT) \cite{pvt} is specially designed for pixel-level dense prediction tasks. Different from ViT and PVT, Swin Transformer \cite{swin} applies self-attention inside the windows and shifts the windows to capture the relation between windows.

What's more, some researchers also want to apply the Transformer structure to object detection. Detection Transformer (DETR) \cite{detr} provides a possible detection paradigm leveraging Transformer which uses the Transformer Decoder to obtain the certain number of bounding boxes. With Hungarian loss, DETR needs no post-processing like Non-Maximum Suppression (NMS).

\begin{figure}[t]
  \centering
  \includegraphics[width=\columnwidth]{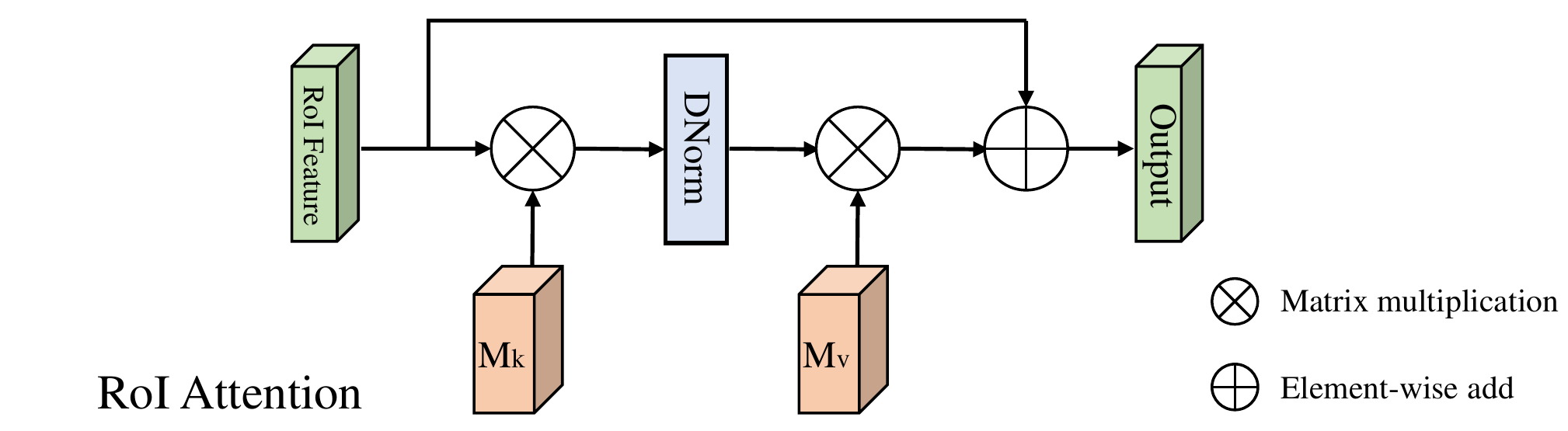} 
  \caption{The illustration of the RoI attention module.}
  \label{fig:roiattn}
\end{figure}

\section{The Proposed Methods}
The overview of the pipeline can be seen in Figure \ref{fig:pipeline}. Our method consists of three modules: RoI attention module, double head, and positional encoding module. 
\subsection{RoI Attention}
A series of works focused on backbone reveals that inter-patch attention can be leveraged to improve the capability of feature extraction. Since RoI can be seen as a patch of original feature maps, it is possible to excavate inter-RoI relationships to boost the feature of RoI features. However, self-attention will generate great costs in computation for the number of RoIs and the features length are huge. External Attention is an efficient substitute for self-attention, which provides a way to capture the relationship between RoIs. We denote it as RoI attention.

\begin{figure}[t]
  \centering
  \includegraphics[width=7cm]{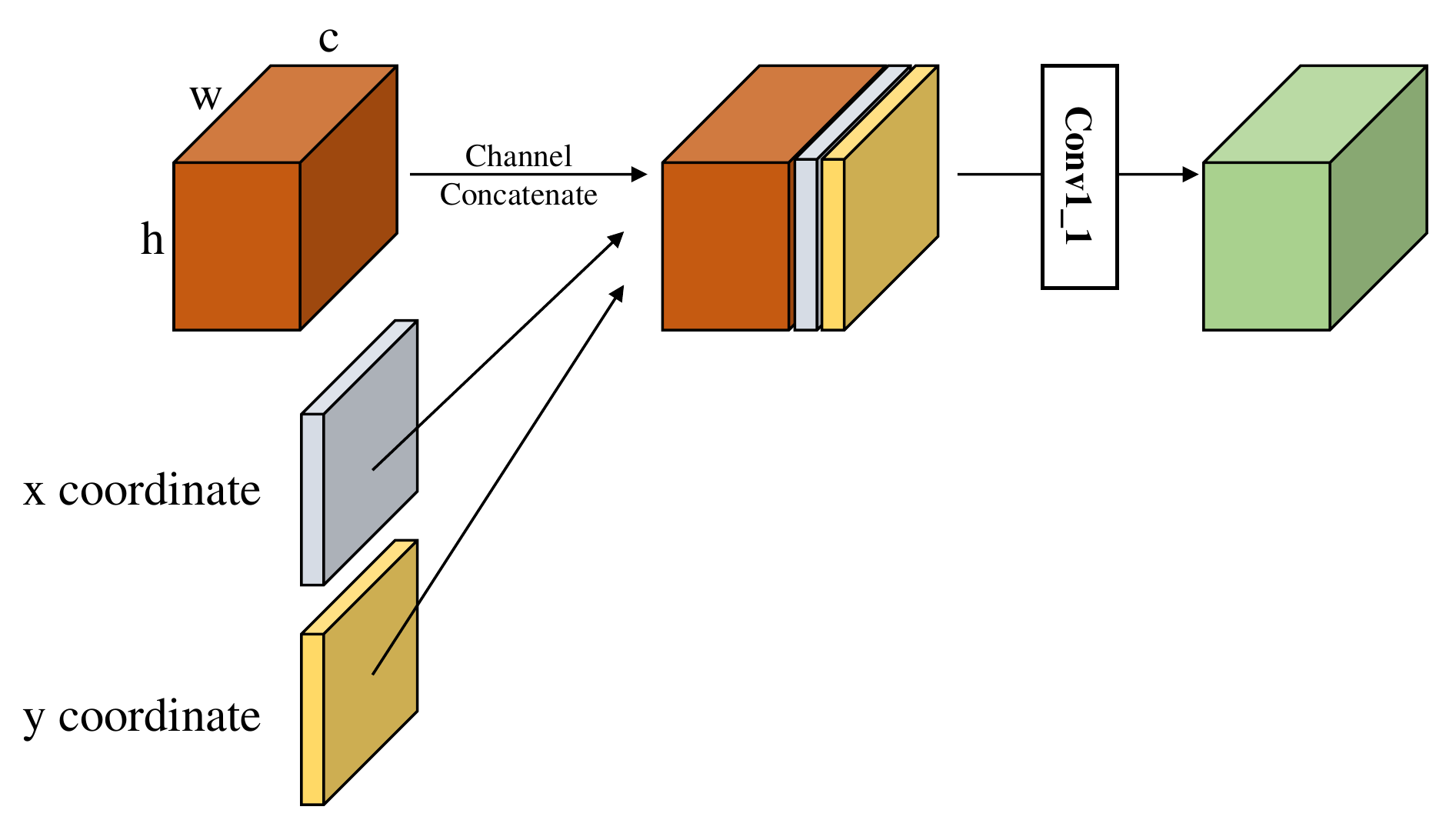} 
  \caption{The illustration of the proposed Positional Encoding module.}
  \label{fig:pos}
\end{figure}

Concretely, a RoI features can be denoted as $X \in \mathbb{R}^{s\times c\times h\times w}$, where $s$ is the number of the RoIs in an image, $c$ is the channels of features, and $h,w$ are the height and width of the RoI features. It can be reshaped as $X \in \mathbb{R}^{s\times L}$ ($L=c \times h \times w$).  As External Attention, we define two external memory units: key memory $M_k \in \mathbb{R}^{L \times d}$ and value memory $M_v \in \mathbb{R}^{L \times d}$. First, we compute the attention between the RoIs and key memory $M_k$:
\begin{equation}
  A=(\alpha)_{i,j}=DNorm(XM_k^T)
\end{equation}
$A$ is the attention maps, and $DNorm$ represents Double Normalization. Different from common Softmax normalization on only one dimension, Double Normalization separately normalizes columns and rows: it uses $Softmax$ to normalize rows and use $L1\_norm$ to normalize columns. In detail, Double Normalization can be formulated as:
\begin{equation}
  \begin{aligned}
  &({\widetilde{\alpha}})_{i,j}=XM_k^T\\
  &\hat{\alpha}_{i,j} = \frac{exp(\widetilde{\alpha}_{i,j})}{\sum_kexp(\widetilde{\alpha}_{k,j})}\\
  &\alpha_{i,j}=\frac{\hat{\alpha}_{i,j}}{\sum_kexp(\hat{\alpha}_{i,k})}
  \end{aligned}
\end{equation}
After Double Normalization, we compute the multiplication of attention maps and value memory to reconstruct the features:
\begin{equation}
  X_{out} =AM_v + X
\end{equation}
Residual connection is leveraged to accelerate the training of the model. In practice, we stack several RoI attention blocks to improve the feature extraction capability.

\begin{table}[t]
  \caption{Comparisons with other object detection methods on UTDAC2020 dataset.}
  \vspace{7pt}
  \centering
  \resizebox{\columnwidth}{!}{
  \begin{tabular}{lcccccc}
    \hline
    Method  & AP & AP50 & AP75 & APS & APM & APL  \\
    \hline
    \multicolumn{2}{l}{\textbf{Two-Stage Detector:}}     &    &    &   &    &  \\
    Faster R-CNN & 44.5  & 80.9  & 44.1  & 20.0  & 39.0  & 50.8  \\
    Double-Head & 45.3  & 81.5  & 45.7  & 20.2  & 40.0  & 51.4  \\
    Libra R-CNN  & 45.8  & 82.0  & 46.4 & 20.1 & 40.2  &  \textbf{52.3} \\
    Dynamic R-CNN & 45.6 & 80.1   & 47.3  & 19.0  & 39.7   & 52.1 \\
    Sparse R-CNN & 45.3  & 81.5  & 45.7  & 20.2  & 40.0  & 51.4  \\
    FPG           & 45.4  & 81.6   & 46.0 & 19.8 & 39.7  &  51.4 \\
    FAFPN       & 45.5  & 82.1  & 45.9 & 18.8 & 39.7  &  51.9 \\
    Groie      & 45.7  & \textbf{82.4}  & 45.6 & 19.9 & 40.1  &  52.0 \\
    \hline
    \multicolumn{2}{l}{\textbf{One-Stage Detector:}}    &    &    &   &    &  \\
    SSD   &  40.0   &  77.5  &  36.5  &  14.7  &  36.1  & 45.1 \\
    RetinaNet & 43.9  &  80.4 & 42.9  & 18.1  & 38.2  &  50.1 \\
    FSAF & 39.7  &  75.8 & 37.9  & 16.8  & 33.6  &  46.4 \\
    NAS-FPN & 37.4  &  70.3 & 35.8  & 12.4  & 36.4  & 40.4 \\
    VFNet & 44.0  &  79.3 & 44.1  & 18.8  & 38.1  & 50.4 \\
    \hline
    \textbf{RoIAttn} & \textbf{46.0}  &  82.0 & \textbf{47.5}  & \textbf{22.9}  & \textbf{40.5}  & 52.2 \\
    \hline
  \end{tabular}
  }
  \label{tab:sota}
\end{table}

\subsection{Double Head}
Object detection consists of two tasks: regression and classification. The author of \cite{doublehead} reveals: for two-stage detectors, the fully connected layer is more suitable for the classification task, while the conv layer is more suitable for the regression task. So a separate branch is needed for these two different tasks. The double head structure is shown in Figure \ref{fig:pipeline}, where the green box denotes the regression branch and the yellow box denotes the classification branch. First, we crop the RoIs from base feature maps for classification branch regression branch respectively with scale factors 1 and 1.3. Second, classification RoI features and regression RoI features are fed into the shared RoI attention module which is mentioned above. Third, for the classification branch, we flatten the RoI features and use two linear layers to obtain classification scores; for the regression branch, four residual bottlenecks are leveraged to extract the regression features, followed by an average pooling and a linear layer to obtain the bounding box's coordinates. For the convolution details of reg branch, we use four layers of convolution, where two layers are used as a block, and two layers are convolved with 3$\times$3$\times$256 and 1$\times$1$\times$1024, respectively, and identity is 1$\times$1$\times$1024.

\subsection{Positional Encoding}
Since regression is a position-sensitive task. CoordConv \cite{coordconv} suggests that it is difficult for convolution to directly regress the coordinate information. Therefore, we think it is logical to add explicit position information to features to improve the regression performance. So we design a positional encoding module (shown in Figure \ref{fig:pos}). And two coordinate maps $C_{x},C_{y} \in \mathbb{R}^{H,W}$:
\begin{equation}
  \begin{aligned}
    &C_x[i,j]=i/W\\
    &C_y[i,j]=j/H
  \end{aligned}
\end{equation}
For the base features $X \in \mathbb{R}^{C \times H \times W}$ from backbone FPN, we concatenate two coordinate maps with base features, and use an $1 \times 1$ conv to down-sample the feature back to original size:
\begin{equation}
  \begin{aligned}
    &X'=concat(X,C_x,C_y), \ \ X' \in \mathbb{R}^{(C+2) \times H \times W}\\
    &X_r = Conv_{1\times 1}(X')
  \end{aligned}
\end{equation}
$X_r$ is the positional encoded features. For different scale features from backbone FPN, we use a shared $1 \times 1$ conv for downsampling. Since we employ a double head structure, only features for the regression branch are encoded.

\begin{table}[t]
  \caption{Abalation study of our method.}
  \vspace{7pt}
  \centering
  \resizebox{\columnwidth}{!}{
  \begin{tabular}{lcccccc}
    \hline
    Methods  & AP & AP50 & AP75 & APS & APM & APL  \\
    \hline
    \multicolumn{2}{l}{\textbf{Single Head}}    &    &    &   &    &  \\
    baseline  & 44.5  & 80.9  & 44.1  & 20.0  & 39.0  & 50.8  \\
    + RoI attention  & 45.4  & 82.1  & 45.8  & 20.3  & 39.6  & 51.9  \\
    \hline
    \multicolumn{2}{l}{\textbf{Double Head}}    &    &    &   &    &  \\
    Only cls. & 45.4   & 81.2  & 46.0  & 22.1  & 39.4  & 51.9  \\
    Only reg. & 45.4   & 81.2  & 46.2  & 21.3  & 39.9  & 51.7  \\
    Both     & 45.8   & 82.0  & 46.6  & 21.7  & 40.0  & 52.2  \\
    \textbf{Full} & \textbf{46.0}  &  \textbf{82.0} & \textbf{47.5}  & \textbf{22.9}  & \textbf{40.5}  & \textbf{52.2} \\
    \hline    
  \end{tabular}
  }
  \label{tab:ablation}
\end{table}

\section{Experiments}
\subsection{Datasets}
We conduct experiments on a challenging object detection datasets UTDAC2020 to validate the generalization performance of our method. UTDAC2020 is an underwater dataset and it short for Underwater Target Detection Algorithm Competition 2020. There are 5168 training images and 1293 validation images. It contains 4 classes: echinus, holothurian, starfish, and scallop. The images in the dataset contain 4 resolutions: $3840 \times 2160$, $1920 \times 1080$, $720 \times 405$, and $586 \times 480$. 

\subsection{Implementation Details}
Our method is implemented on MMdetection. We adopt FPN-based Faster R-CNN with ResNet-50 model pre-trained on ImageNet as our baseline. The models are trained for 12 epochs with an initial learning rate of 0.005, which will be decreased by 0.1 after 8 and 11 epochs respectively. Our method is trained on a single NVIDIA V100 GPU, with batch size set to 4. SGD is adapted as an optimizer, where the weight decay is 0.0001 and the momentum is 0.9. In the experiments, no data augmentation except for the traditional horizontal flipping is utilized. All other hyper-parameters follow the settings in MMdetection.

\subsection{Experiments on UTDAC2020 Dataset}
We compare our proposed method with several state-of-the-art object detection methods, including two-stage detectors: Faster R-CNN \cite{ren2015faster}, Double-Head \cite{doublehead}, Libra R-CNN \cite{pang2019libra}, Dynamic R-CNN \cite{zhang2020dynamic}, Sparse R-CNN \cite{sun2021sparse}, FPG \cite{fpg}, PA-FPN \cite{pafpn}, GRoIE \cite{groie} and one-stage detector: SSD \cite{liu2016ssd}, RetinaNet \cite{retinanet}, FSAF \cite{fsaf}, NAS-FPN \cite{nasfpn} and VFNet \cite{zhang2020varifocalnet}. For fair comparison, all detectors are based on ResNet-50+FPN paradigm. The results show that our method achieve 46.0\% mAP performance, 1.5\% higher than the baseline (Faster R-CNN) and equal to Libra R-CNN. 


\begin{table}[ht]
  \caption{Abalation study of $d$ and the depth of RoI attention.}
  \vspace{7pt}
  \centering
  \resizebox{\columnwidth}{!}{
  \begin{tabular}{cccccccc}
    \hline
    $d$  & $depth$  & AP & AP50 & AP75 & APS & APM & APL  \\
    \hline
    10  & 1 & \textbf{45.4}   & 82.1  & 45.8  & 20.3  & 39.6  & \textbf{51.9} \\
    20  & 1 & 45.1   & 81.0  & 46.0  & 19.3  & \textbf{39.7}  & 51.1  \\
    40  & 1 & 45.1   & 81.0  & 46.0  & 19.3  & \textbf{39.7}  & 51.1  \\
    80  & 1 & 45.0   & 81.7  & 45.1  & 19.1  & 39.2  & 51.4  \\
    \hline
    10  & 2 & 44.9   & 81.6  & 44.6  & 22.1  & 39.4  & 50.9  \\
    20  & 2 & 45.3  & 81.6   & \textbf{46.1}  & 19.1  & 39.4  & 51.7 \\
    40  & 2 & 45.0   & 81.0  & \textbf{46.1}  & 21.0  & 39.2  & 51.5  \\
    80  & 2 & 45.2  & 82.0    & 45.3  & 20.3  & 39.4  &  51.5 \\
    \hline
    10  & 3 & 45.1   & 81.5  & 45.3  & 19.8  & 39.4  & 51.4 \\
    20  & 3 & 45.0   & 81.7  & 45.1  & 23.2  & 39.1  & 51.5  \\
    40  & 3 & 44.9   & 81.9  & 44.6  & 20.5 & 39.5  & 51.1  \\
    80  & 3 & 45.1   & \textbf{82.2}  & 44.9  & \textbf{20.8}  & 39.3  & 51.5  \\
    \hline    
  \end{tabular}
  }
  \label{tab:depth and dimension}
\end{table}

\subsection{PASCAL VOC Dataset}

The experiment results are shown in Table \ref{tab:psacal voc}. RoIAttn can also achieve satisfactory results in PASCAL VOC Dataset. Our method achieves an improvement of +1.6 mAP over the baseline model. In addition, our method achieves a +1.1mAP higher than Cascade R-CNN. It is worth noting that our settings on the Pascal VOC Dataset are slightly different from those on the UTDAC2020 Dataset. Specifically, to better meet the characterization needs of the 20 different categories in Pascal VOC Dataset, we set the d and depth in Table \ref{tab:depth and dimension} to 20 and 2.

\begin{table}[ht]
  \caption{Comparisons with other object detection methods on PASCAL VOC dataset.}
  \vspace{7pt}
  \centering
  \resizebox{\columnwidth}{!}{
  \begin{tabular}{lccc}
    \hline
    Method   & Backbone & Input Size & mAP \\
    \hline
    Faster R-CNN w/FPN &   ResNet50  &  1000 $\times$ 600  &  79.5 \\
    RetinaNet w/FPN &  ResNet50  &  1000 $\times$ 600 & 77.3 \\
    SSD300    &  VGG16  &  300 $\times$ 300 & 74.0 \\
    SSD512    &  VGG16  &  512 $\times$ 512 & 75.3 \\
    Cascade R-CNN & ResNet50 & 1000 $\times$ 600  &  80.0 \\
    \hline
    \textbf{RoIAttn} &  ResNet50  &  1000 $\times$ 600 & \textbf{81.1} \\
    \hline
  \end{tabular}
  }
  \label{tab:psacal voc}
\end{table}

\subsection{Ablation Studies}
In this section, we perform a series of ablation experiments on the various hyper-parameters of our method on the UTDAC2020 dataset to validate its effectiveness.

First of all, we analyze the effectiveness of our method (shown in Table \ref{tab:ablation}). When we apply RoI attention module on our baseline (44.5\% mAP), the performance increase by 0.8\% mAP. When we change the R-CNN head from the single head structure to the double head structure, and apply the RoI attention module on both branches, the performance will reach 45.8\% mAP, because double head structure can decouple the regression and classification task to increase the performance. The performance will decline if we only apply RoI attention in classification branch (Only cls.) and only regression branch (Only reg.). 
If we further add the positional encoding module to the regression branch, the performance will further increase to 46.0\% mAP. Positional encoding applying on Regression branch give a direct positional information to feature maps, providing accurate regressed boxes. (``Full'' has the same AP50 with Both, but achieves higher performance in AP and AP75.) 
The experiments indicate that our proposed RoI attention is effective for extracting the salient feature from the RoI samples.

Next, we will analyze how depth and the hidden dimension $d$ in $M_k$ and $M_v$ affect the performance. In this experiment, no double head and positional encoding are used. As is shown in Table \ref{tab:depth and dimension}, we can conclude that simply increasing the depth of the hidden dimension of the attention module does not guarantee any increase in the performance. Stacked RoI attention can not improve the performance, because RoI attention can be seen as a clustering methods, and one module is enough for convergence. The d in table 3 can be seen as the number of the clustering center. Because the classes in the UTDAC2020 dataset only contains 4 classes, d = 10 is enough for representing the semantic information. The best performance is 45.4\% mAP and is obtained when $d$ is 10 and the depth is 1.

\section{Conclusion}
In this paper, we propose a brand new object detector RoIAttn, which consists of three components: RoI attention module, double head, and positional encoding module. RoI attention module can capture the relation of RoIs and enhance the features. Double head structure leverages task-specific characteristics, and designs different branches for different tasks. The positional Encoding module helps the detectors to learn coordinate information. The experiments on UTDAC2020 show its effectiveness on underwater object detection.

{
\bibliographystyle{IEEEbib}
\bibliography{refs}

\begin{thebibliography}{10}

\bibitem{Bert}
Jacob Devlin, Ming-Wei Chang, Kenton Lee, and Kristina Toutanova,
\newblock ``Bert: Pre-training of deep bidirectional transformers for language
  understanding,''
\newblock {\em arXiv preprint arXiv:1810.04805}, 2018.

\bibitem{gpt1}
Alec Radford, Karthik Narasimhan, Tim Salimans, and Ilya Sutskever,
\newblock ``Improving language understanding by generative pre-training,''
\newblock {\em OpenAI blog}, 2018.

\bibitem{gpt2}
Alec Radford, Jeffrey Wu, Rewon Child, David Luan, Dario Amodei, Ilya
  Sutskever, et~al.,
\newblock ``Language models are unsupervised multitask learners,''
\newblock {\em OpenAI blog}, vol. 1, no. 8, pp. 9, 2019.

\bibitem{Roberta}
Yinhan Liu, Myle Ott, Naman Goyal, Jingfei Du, Mandar Joshi, Danqi Chen, Omer
  Levy, Mike Lewis, Luke Zettlemoyer, and Veselin Stoyanov,
\newblock ``Roberta: A robustly optimized bert pretraining approach,''
\newblock {\em arXiv preprint arXiv:1907.11692}, 2019.

\bibitem{Transformer}
Ashish Vaswani, Noam Shazeer, Niki Parmar, Jakob Uszkoreit, Llion Jones,
  Aidan~N Gomez, {\L}ukasz Kaiser, and Illia Polosukhin,
\newblock ``Attention is all you need,''
\newblock in {\em Advances in neural information processing systems}, 2017, pp.
  5998--6008.

\bibitem{nl}
Xiaolong Wang, Ross Girshick, Abhinav Gupta, and Kaiming He,
\newblock ``Non-local neural networks,''
\newblock in {\em Proceedings of the IEEE conference on computer vision and
  pattern recognition}, 2018, pp. 7794--7803.

\bibitem{crisscross}
Zilong Huang, Xinggang Wang, Lichao Huang, Chang Huang, Yunchao Wei, and Wenyu
  Liu,
\newblock ``Ccnet: Criss-cross attention for semantic segmentation,''
\newblock in {\em Proceedings of the IEEE/CVF International Conference on
  Computer Vision}, 2019, pp. 603--612.

\bibitem{standalone}
Prajit Ramachandran, Niki Parmar, Ashish Vaswani, Irwan Bello, Anselm Levskaya,
  and Jon Shlens,
\newblock ``Stand-alone self-attention in vision models,''
\newblock {\em Advances in Neural Information Processing Systems}, vol. 32,
  2019.

\bibitem{ea}
Meng-Hao Guo, Zheng-Ning Liu, Tai-Jiang Mu, and Shi-Min Hu,
\newblock ``Beyond self-attention: External attention using two linear layers
  for visual tasks,''
\newblock {\em arXiv preprint arXiv:2105.02358}, 2021.

\bibitem{ViT}
Alexey Dosovitskiy, Lucas Beyer, Alexander Kolesnikov, Dirk Weissenborn,
  Xiaohua Zhai, Thomas Unterthiner, Mostafa Dehghani, Matthias Minderer, Georg
  Heigold, Sylvain Gelly, et~al.,
\newblock ``An image is worth 16x16 words: Transformers for image recognition
  at scale,''
\newblock in {\em International Conference on Learning Representations}, 2020.

\bibitem{swin}
Ze~Liu, Yutong Lin, Yue Cao, Han Hu, Yixuan Wei, Zheng Zhang, Stephen Lin, and
  Baining Guo,
\newblock ``Swin transformer: Hierarchical vision transformer using shifted
  windows,''
\newblock in {\em IEEE International Conference on Computer Vision}, 2021.

\bibitem{pvt}
Wenhai Wang, Enze Xie, Xiang Li, Deng-Ping Fan, Kaitao Song, Ding Liang, Tong
  Lu, Ping Luo, and Ling Shao,
\newblock ``Pyramid vision transformer: A versatile backbone for dense
  prediction without convolutions,''
\newblock in {\em IEEE International Conference on Computer Vision}, 2021.

\bibitem{deit}
Hugo Touvron, Matthieu Cord, Matthijs Douze, Francisco Massa, Alexandre
  Sablayrolles, and Herv{\'e} J{\'e}gou,
\newblock ``Training data-efficient image transformers \& distillation through
  attention,''
\newblock in {\em International Conference on Machine Learning}. PMLR, 2021,
  pp. 10347--10357.

\bibitem{ren2015faster}
Shaoqing Ren, Kaiming He, Ross Girshick, and Jian Sun,
\newblock ``Faster r-cnn: Towards real-time object detection with region
  proposal networks,''
\newblock {\em Advances in neural information processing systems}, vol. 28, pp.
  91--99, 2015.

\bibitem{sun2021sparse}
Peize Sun, Rufeng Zhang, Yi~Jiang, Tao Kong, Chenfeng Xu, Wei Zhan, Masayoshi
  Tomizuka, Lei Li, Zehuan Yuan, Changhu Wang, et~al.,
\newblock ``Sparse r-cnn: End-to-end object detection with learnable
  proposals,''
\newblock in {\em Proceedings of the IEEE/CVF Conference on Computer Vision and
  Pattern Recognition}, 2021, pp. 14454--14463.

\bibitem{doublehead}
Yue Wu, Yinpeng Chen, Lu~Yuan, Zicheng Liu, Lijuan Wang, Hongzhi Li, and Yun
  Fu,
\newblock ``Rethinking classification and localization for object detection,''
\newblock in {\em Proceedings of the IEEE/CVF conference on computer vision and
  pattern recognition}, 2020, pp. 10186--10195.

\bibitem{cai2018cascade}
Zhaowei Cai and Nuno Vasconcelos,
\newblock ``Cascade r-cnn: Delving into high quality object detection,''
\newblock in {\em Proceedings of the IEEE conference on computer vision and
  pattern recognition}, 2018, pp. 6154--6162.

\bibitem{zhang2020dynamic}
Hongkai Zhang, Hong Chang, Bingpeng Ma, Naiyan Wang, and Xilin Chen,
\newblock ``Dynamic r-cnn: Towards high quality object detection via dynamic
  training,''
\newblock in {\em European Conference on Computer Vision}. Springer, 2020, pp.
  260--275.

\bibitem{liu2016ssd}
Wei Liu, Dragomir Anguelov, Dumitru Erhan, Christian Szegedy, Scott Reed,
  Cheng-Yang Fu, and Alexander~C Berg,
\newblock ``Ssd: Single shot multibox detector,''
\newblock in {\em European conference on computer vision}. Springer, 2016, pp.
  21--37.

\bibitem{retinanet}
Tsung-Yi Lin, Priya Goyal, Ross Girshick, Kaiming He, and Piotr Doll{\'a}r,
\newblock ``Focal loss for dense object detection,''
\newblock in {\em Proceedings of the IEEE international conference on computer
  vision}, 2017, pp. 2980--2988.

\bibitem{fsaf}
Chenchen Zhu, Yihui He, and Marios Savvides,
\newblock ``Feature selective anchor-free module for single-shot object
  detection,''
\newblock in {\em Proceedings of the IEEE/CVF Conference on Computer Vision and
  Pattern Recognition}, 2019, pp. 840--849.

\bibitem{detr}
Nicolas Carion, Francisco Massa, Gabriel Synnaeve, Nicolas Usunier, Alexander
  Kirillov, and Sergey Zagoruyko,
\newblock ``End-to-end object detection with transformers,''
\newblock in {\em European Conference on Computer Vision}. Springer, 2020, pp.
  213--229.

\bibitem{coordconv}
Rosanne Liu, Joel Lehman, Piero Molino, Felipe~Petroski Such, Eric Frank, Alex
  Sergeev, and Jason Yosinski,
\newblock ``An intriguing failing of convolutional neural networks and the
  coordconv solution,''
\newblock {\em arXiv preprint arXiv:1807.03247}, 2018.

\bibitem{pang2019libra}
Jiangmiao Pang, Kai Chen, Jianping Shi, Huajun Feng, Wanli Ouyang, and Dahua
  Lin,
\newblock ``Libra r-cnn: Towards balanced learning for object detection,''
\newblock in {\em Proceedings of the IEEE/CVF Conference on Computer Vision and
  Pattern Recognition}, 2019, pp. 821--830.

\bibitem{fpg}
Kai Chen, Yuhang Cao, Chen~Change Loy, Dahua Lin, and Christoph Feichtenhofer,
\newblock ``Feature pyramid grids,''
\newblock {\em arXiv preprint arXiv:2004.03580}, 2020.

\bibitem{pafpn}
Shu Liu, Lu~Qi, Haifang Qin, Jianping Shi, and Jiaya Jia,
\newblock ``Path aggregation network for instance segmentation,''
\newblock in {\em Proceedings of IEEE Conference on Computer Vision and Pattern
  Recognition (CVPR)}, 2018.

\bibitem{groie}
Leonardo Rossi, Akbar Karimi, and Andrea Prati,
\newblock ``A novel region of interest extraction layer for instance
  segmentation,'' 2020.

\bibitem{nasfpn}
Golnaz Ghiasi, Tsung-Yi Lin, and Quoc~V Le,
\newblock ``Nas-fpn: Learning scalable feature pyramid architecture for object
  detection,''
\newblock in {\em Proceedings of the IEEE/CVF Conference on Computer Vision and
  Pattern Recognition}, 2019, pp. 7036--7045.

\bibitem{zhang2020varifocalnet}
Haoyang Zhang, Ying Wang, Feras Dayoub, and Niko S{\"u}nderhauf,
\newblock ``Varifocalnet: An iou-aware dense object detector,''
\newblock {\em arXiv preprint arXiv:2008.13367}, 2020.

\end{thebibliography}
}
\end{document}